\documentclass[10pt,twocolumn,letterpaper]{article}

\usepackage{cvpr}
\usepackage{times}
\usepackage{epsfig}
\usepackage{graphicx}
\usepackage{amsmath}
\usepackage{amssymb}
\usepackage{booktabs}
\usepackage{subcaption}
\usepackage{stackengine}
\usepackage{epstopdf}
\usepackage{float}
\usepackage{colortbl}
\usepackage{color}

\newcommand{\norm}[1]{\left\lVert#1\right\rVert}

\definecolor{Yellow}{rgb}{1,1, 0.7}

\usepackage[breaklinks=true,bookmarks=false]{hyperref}

\cvprfinalcopy 

\def\cvprPaperID{3761} 

\ifcvprfinal\fi
\begin{document}

\title{Burst Denoising with Kernel Prediction Networks}


\newcommand{\newand}{\quad\quad\,\,}

\author{
Ben Mildenhall$^{1,2^*}$\phantom{\thanks{Work done while interning at Google.}}
\newand
Jonathan T. Barron$^2$
\newand
Jiawen Chen$^2$ \\
Dillon Sharlet$^2$
\newand
Ren Ng$^1$
\newand
Robert Carroll$^2$ \\
$^1$UC Berkeley \newand $^2$Google Research
}

\maketitle

\thispagestyle{empty}

\begin{abstract}
We present a technique for jointly denoising bursts of images taken from a handheld camera.
In particular, we propose a convolutional neural network architecture for predicting spatially varying kernels that can both align and denoise frames, a synthetic data generation approach based on a realistic noise formation model, and an optimization guided by an annealed loss function to avoid undesirable local minima. Our model matches or outperforms the state-of-the-art across a wide range of noise levels on both real and synthetic data.

\end{abstract}


\section{Introduction}

\begin{figure*}
\begin{center}
\includegraphics[width=\textwidth]{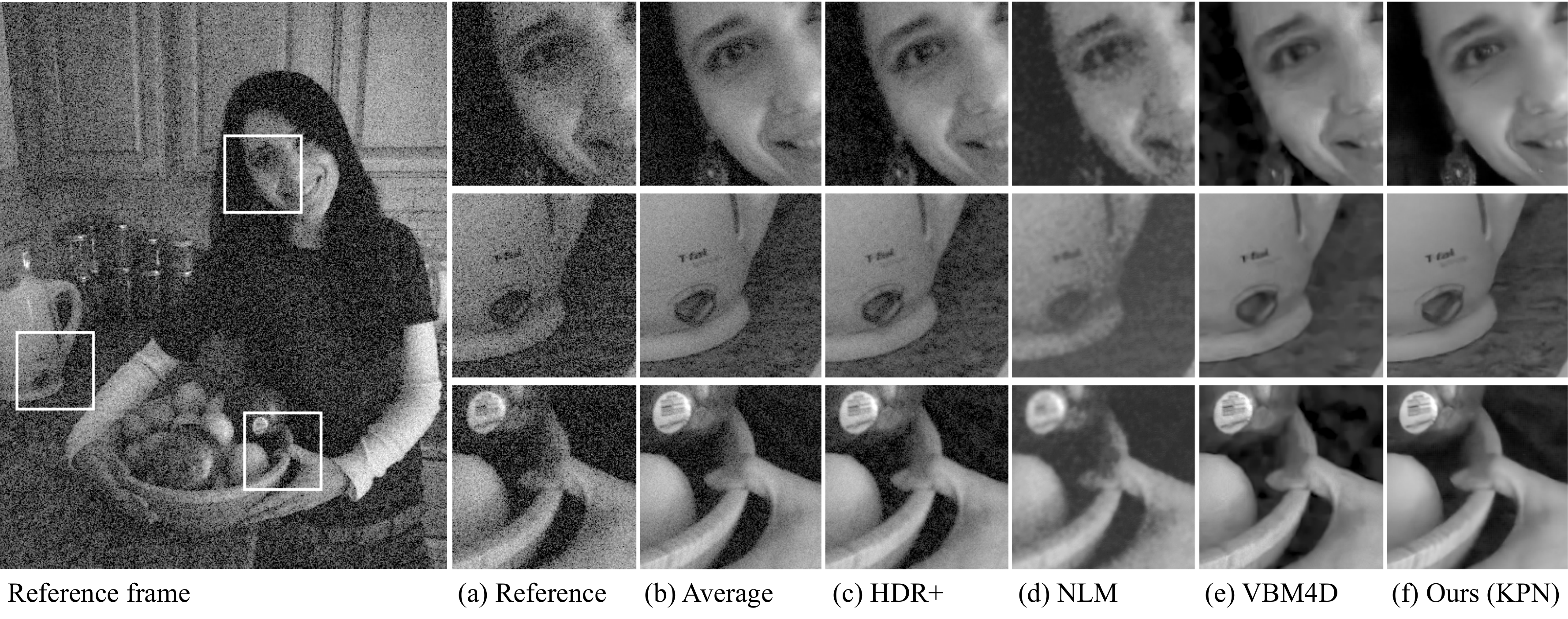}
\caption{
A qualitative evaluation of our model on real image bursts from a handheld camera in a low-light environment.
The reference frame from the input burst (a) is sharp, but noisy.
Noise can be reduced by simply averaging a burst of similar images (b), but this can fail in the presence of motion (see Figure~\ref{fig:pianoman}).
Our approach (f) learns to use the information present in the entire burst to denoise a single frame, producing lower noise and avoiding artifacts compared to baseline techniques (c -- e). See the supplement for full resolution images and more examples.
 \label{fig:realtable1}
 }
\end{center}
\end{figure*}

The task of image denoising is foundational to the study of imaging and computer vision.
Traditionally, the problem of single-image denoising has been addressed as one of statistical inference using analytical priors~\cite{PeronaMalik1990, Rudin1992}, but recent work has built on the success of deep learning by using convolutional neural networks that learn mappings from noisy images to noiseless images by training on millions of examples~\cite{ResidDenoising}.
These networks appear to learn the likely appearance of ``ground truth'' noiseless images in addition to the statistical properties of the noise present in the input images.

Multiple-image denoising has also traditionally been approached through the lens of classical statistical inference, under the assumption that averaging multiple noisy and independent samples of a signal will result in a more accurate estimate of the true underlying signal.
However, when denoising image bursts taken with handheld cameras, simple temporal averaging yields poor results because of scene and camera motion. Many techniques attempt to first align the burst or include some notion of translation-invariance within the denoising operator itself~\cite{HdrPlus}.
The idea of denoising by combining multiple aligned image patches is also key to many of the most successful single image techniques~\cite{NonlocalMeans, BM3D}, which rely on the self-similarity of a single image to allow some degree of denoising via averaging.

We propose a method for burst denoising with the signal-to-noise ratio benefits of multi-image denoising and the large capacity and generality of convolutional neural networks.
Our model is capable of matching or outperforming the state-of-the-art at all noise levels on both synthetic and real data.
Our contributions include:
\begin{enumerate}
  \setlength{\itemsep}{1pt}
  \setlength{\parskip}{0pt}
  \setlength{\parsep}{10pt}
\item A procedure for converting post-processed images taken from the internet into data with the characteristics of raw linear data captured by real cameras. This lets us to train a model that generalizes to real images and circumvents the difficulties in acquiring ground truth data for our task from a camera.
\item A network architecture that outperforms the state-of-the-art on synthetic and real data by predicting a unique 3D denoising kernel to produce each pixel of the output image. This provides both a performance improvement over a network that synthesizes pixels directly, and a way to visually inspect how each burst image is being used. 
\item A training procedure for our kernel prediction network that allows it to predict filter kernels that use information from multiple images even in the presence of small unknown misalignments.
\item A demonstration that a network that takes the noise level of the input imagery as input during training and testing generalizes to a much wider range of noise levels than a blind denoising network.
\end{enumerate}

\section{Related work}

Single-image denoising is a longstanding problem, originating with classical methods like anisotropic diffusion~\cite{PeronaMalik1990} or total variation denoising~\cite{Rudin1992}, which used analytical priors and non-linear optimization to recover a signal from a noisy image.
These ideas were built upon to develop multi-image or video denoising techniques such as VBM4D~\cite{VBM4D} and non-local means~\cite{NonlocalMeans,ReliableMotion}, which group similar patches across time and jointly filter them under the assumption that multiple noisy observations can be averaged to better estimate the true underlying signal.
Recently these ideas have been retargeted towards the task of denoising a burst of noisy images captured from commodity mobile phones, with an emphasis on energy efficiency and speed~\cite{HdrPlus,FastBurst}. These approaches first align image patches to within a few pixels and then perform joint denoising by robust averaging (such as Wiener filtering). Another line of work has focused on achieving high quality by combining multiple image formation steps with a single linear operator and using modern optimization techniques to solve the associated inverse problem~\cite{FlexIsp,Proximal}. These approaches generalize to multiple image denoising but require calculating alignment as part of the forward model.

The success of deep learning has yielded a number of neural network approaches to multi-image denoising~\cite{ResidDenoising, DeepRNNs}, in addition to a wide range of similar tasks such as joint denoising and demosaicking~\cite{GharbiDemosaic}, deblurring~\cite{DeepDeblurring}, and superresolution~\cite{DetailRevealing}.
Similar in spirit to our method, Kernel-Predicting Networks~\cite{KernelPredicting} denoise Monte Carlo renderings with a network that generates a filter for every pixel in the desired output, which constrains the output space and thereby prevents artifacts.
Similar ideas have been applied successfully to both video interpolation~\cite{AdaConv, AdaSepConv} and video prediction~\cite{PhysInteract, VoxelFlow, CrossConv, DynamicFilter}, where applying predicted optical flow vectors or filters to the input image data helps prevent the blurry outputs often produced by direct pixel synthesis networks.

\section{Problem specification}

Our goal is to produce a single clean image from a noisy burst of $N$ images captured by a handheld camera.
Following the design of recent work~\cite{HdrPlus}, we select one image $X_1$ in the burst as the ``reference'' and denoise it with the help of ``alternate'' frames $X_2,\ldots,X_N$. It is not necessary for $X_1$ to be the first image acquired.
All input images are in the raw linear domain to avoid losing signal due to the post-processing performed between capture and display (e.g., demosaicking, sharpening, tone mapping, and compression).
Creating training examples for this task requires careful consideration of the characteristics of raw sensor data.

\subsection{Characteristics of raw sensor data}

Camera sensors output raw data in a linear color space, where pixel measurements are proportional to the number of photoelectrons collected.
The primary sources of noise are shot noise, a Poisson process with variance equal to the signal level, and read noise, an approximately Gaussian process caused by a variety of sensor readout effects.
These effects are well-modeled by a signal-dependent Gaussian distribution~\cite{Healey1994}:
\begin{equation}
x_p \sim \mathcal N \left(y_p, \sigma_r^2 + \sigma_s y_p \right) \label{eq:srnoise}
\end{equation}
where $x_p$ is a noisy measurement of the true intensity $y_p$ at pixel $p$. The noise parameters $\sigma_r$ and $\sigma_s$ are fixed for each image but can vary across images as sensor gain (ISO) changes\footnote{These noise parameters are part of the Adobe DNG specification. Most cameras are calibrated to output them as a function of analog and digital gain.}.

The sensor outputs pixel measurements in the integer-quantized range $[0,2^B)$, where $B$ is the sensor's bit depth. Clipping against the upper end of the range can be avoided by underexposing the photo. The sensor itself avoids clipping potentially negative read noise values against zero by adding a constant positive offset called the ``black level'' to every pixel before measurement. This offset must be subtracted in order to make sure that the expected value of a completely black pixel is truly zero.

Real image bursts contain motion from both hand shake and scene motion. Hand motion can often be well estimated with a global model, while scene motion requires local estimation. Motion may cause disocclusions, thereby rendering accurate correspondence impossible.

\subsection{Synthetic training data}

Gathering ground truth data for image restoration tasks is challenging, as it is constrained by the maximum performance of the imaging system---it is unlikely that we can learn to denoise beyond the quality of the ground truth examples.
Plotz et al.~\cite{BenchmarkingDenoising} describe the many issues with creating a ground truth dataset for single-image denoising.
Burst denoising adds an additional complication since methods must be robust to some degree of misalignment between the images. 
Because deep neural networks require millions of image patches during training, it is impractical to use real pairs of noisy and noise-free ground truth bursts.
We therefore synthesize training data, using images from the Open Images dataset~\cite{Openimages}.
These images are modified to introduce synthetic misalignment and noise approximating the characteristics of real image bursts.

To generate a synthetic burst of $N$ frames, we take a single image and generate $N$ cropped patches with misalignments ${\Delta_i}$, where each $\Delta_i$ is drawn from a 2D uniform integer distribution.
We downsample these patches by $J=4$ in each dimension using a box filter, which reduces noise and compression artifacts.
We constrain our random crops such that after downsampling, the alternate frames have a maximum translation of $\pm 2$ pixels relative to the reference. 

It is critical to also simulate complete alignment failure to provide robustness in the presence of occlusion or large scene motion. Some real bursts will be easy to align and some hard, so for each burst we pick an approximate number of misaligned frames $n\sim \mathrm{Poisson}(\lambda)$. Then for each alternate frame in that burst, we sample a coin flip with probability $n/N$ to decide whether to apply a translational shift of up to $\pm 16$ pixels after downsampling relative to the reference. For synthetic bursts of length 8, we use $\lambda=1.5$. We expect our model to correct for the $\pm 2$ pixel misalignment but not for the $\pm 16$ pixel misalignment.

\begin{figure}
\begin{center}
  \includegraphics[width=1\linewidth]{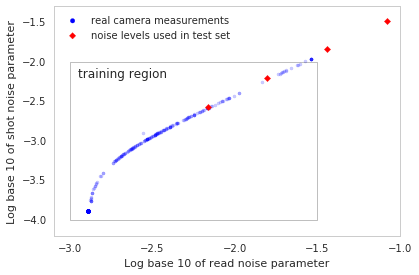}
\end{center}
\vspace{-0.2in}
  \caption{Shot and read noise parameters are tightly coupled for a digital camera sensor. In blue, we show shot/read parameter pairs from hundreds of images taken with the same cellphone camera. In red, we show the shot and read values corresponding to the synthetic gain levels we use for evaluation in Table~\ref{table:errA}. During training we sample shot and read values uniformly at random from the entire rectangular area identified here, as different camera sensors may trace out different shot versus read noise curves.}
\label{fig:gains}
\end{figure}

To generate synthetic noise, we first invert gamma correction on our collection of randomly perturbed and downsampled crops to yield a set of patches in an approximately linear color space. Then, we linearly scale the data by a value randomly sampled from $[0.1, 1]$. This compresses the histogram of intensities to more closely match our real data, which is underexposed to avoid highlight clipping as in the HDR+ burst imaging pipeline~\cite{HdrPlus}. Finally, we sample shot and read factors $\sigma_r, \sigma_s$ from ranges that match what we observe in real data (see Fig.~\ref{fig:gains}) and add noise to the burst images by sampling from the distribution of Eq.~\ref{eq:srnoise}.

\section{Model}

\begin{figure*}[t!]
\begin{center}
  \includegraphics[width=1\linewidth]{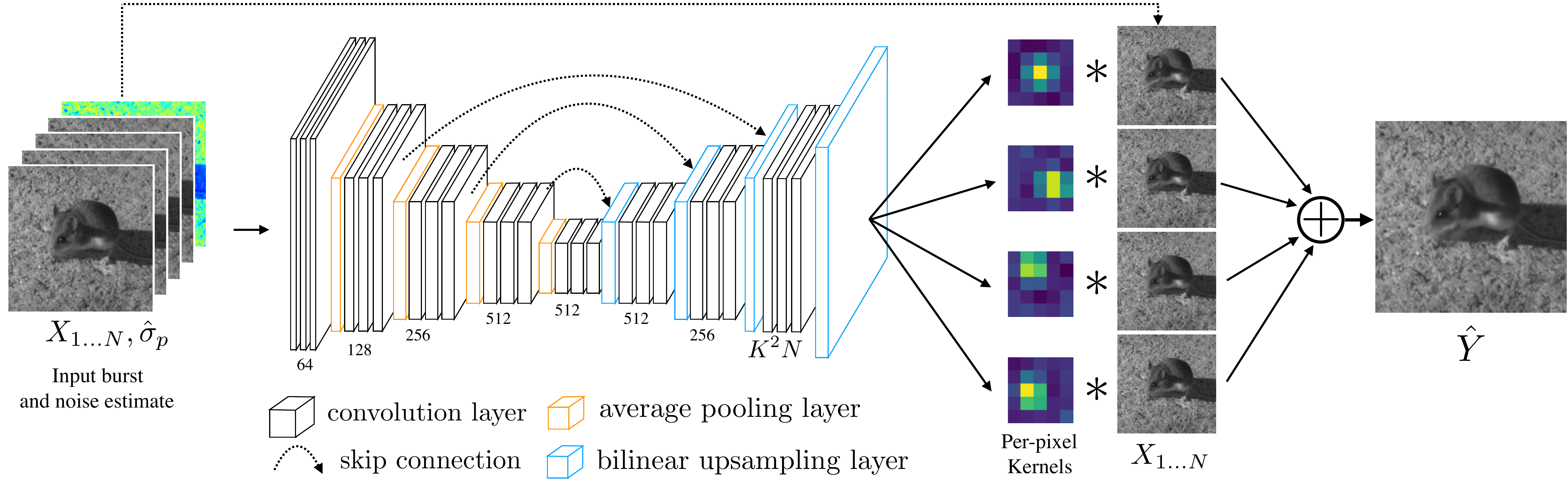}
\end{center}
  \caption{Our KPN architecture for burst denoising is based on the encoder-decoder structure in~\cite{AdaSepConv}, which outputs per-pixel feature vectors. These vectors are then reshaped into a set of spatially-varying kernels that are applied to the input burst.}
\label{fig:architecture}
\end{figure*}

Our model builds upon recent work in kernel prediction networks. Niklaus \etal.~\cite{AdaConv, AdaSepConv} perform video interpolation by generating a stack of two filters at each pixel location, then applying these filter kernels to pairs of input frames. Bako \etal.~\cite{KernelPredicting} use a similar idea to generate per-pixel denoising kernels for specifically for Monte Carlo renderings. Our model combines both of these applications of kernel predicting networks: it generates a stack of per-pixel filter kernels that jointly aligns, averages, and denoises a burst to produce a clean version of the reference frame.

Our kernel prediction network (KPN) uses an encoder-decoder architecture with skip connections closely resembling the architecture in~\cite{AdaSepConv} (see Fig.~\ref{fig:architecture}). Rather than directly synthesizing the pixels of an output image with a single output channel, the KPN has $K^2N$ output channels, which is reshaped into a stack of $N$ $K {\times} K$ linear filters at each pixel. The value at each pixel $p$ in our output $\hat Y$ is
\begin{equation}
\hat Y^p = \frac 1 N \sum_{i=1}^N \big\langle f_i^p, V^p(X_i) \big\rangle,
\end{equation}
where $V^p(X_i)$ is the $K \times K$ neighborhood of pixel $p$ in image $X_i$ and $f_i^p$ is its corresponding kernel.
$\hat Y$ is the result of applying a spatially varying kernel to each image (a dot product) then computing the mean over time. We will also use the shorthand $\hat Y = \frac 1 N \sum_{i=1}^N f_i(X_i)$ to denote computing the two dimensional output image as a whole. In our experiments, $K=5$ and $N=8$.

In addition to the raw burst, the network takes a per-pixel estimate of the standard deviation of the signal as input, similar to Gharbi et al.~\cite{GharbiDemosaic}. We estimate the noise at each pixel $p$ to be
\begin{equation}
\hat\sigma_p = \sqrt{\sigma_r^2 + \sigma_s \max(x_p, 0) }
\label{eq:sigmap}
\end{equation}
where $x_p$ is the intensity of pixel $p$ in the first image of the burst.
This noise estimate is necessarily approximate because we are substituting the observed intensity $x_p$ for the true intensity $y_p$. We assume $\sigma_r$ and $\sigma_s$ are known. The benefits and tradeoffs of providing the noise level to the network are discussed in Section~\ref{nonblind}.

Unlike Bako et al.~\cite{KernelPredicting} we do not normalize the predicted filters with a softmax, thereby allowing predicted kernels to have negative values. We also found softmax normalization to lead to unstable gradients during training.

\subsection{Basic loss function}

Our basic loss is a weighted average of $L^2$ distance on pixel intensities and $L^1$ distance on pixel gradients as compared to the ground truth image.
We apply the loss after restoring the white level to $1$ and applying the sRGB transfer function for gamma correction, which produces a more perceptually relevant estimate.
Computing the loss without gamma correction overemphasizes errors in the highlights and produces overly blurry or patchy shadows.

Our basic loss on an output image patch $\hat Y$ and its ground truth image patch $Y^*$ is
\begin{equation}
\resizebox{1.0\linewidth}{!}{$
\displaystyle
\ell(\hat Y, Y^*) = \lambda_2 \norm{\Gamma (\hat Y) - \Gamma(Y^*)}_2^2 + \lambda_1 \norm{\nabla \Gamma(\hat Y) - \nabla \Gamma(Y^*)}_1.
$}
\end{equation}
Here $\nabla$ is the finite difference operator that convolves its input with $[-1,1]$ and $[-1,1]^\mathrm{T}$, and $\lambda_2$ and $\lambda_1$ are fixed constants (both set to 1 in our experiments). $\Gamma$ is the sRGB transfer function~\cite{srgb}:
\begin{align}
\Gamma(X) & = {\begin{cases}
12.92 X, & X\leq 0.0031308 \\
(1+a)X^{1/2.4}-a, & X > 0.0031308\end{cases}} \nonumber \\
a & =0.055
\end{align}
This choice of transfer function was necessary for successful gradient-based optimization. We could not simply apply the straightforward gamma correction function $X^\gamma$ because its gradient approaches infinity as $X$ approaches $0$ (which can cause exploding gradients during optimization) and is undefined for negative values of $X$ (which we encounter throughout training due to the negative values in the input after black level subtraction, and because the sign of model output is unconstrained).

\subsection{Annealed loss term}
\label{sec:annealed}

Minimizing our loss $\ell(\hat Y, Y^*)$ with respect to the KPN model weights is straightforward, as our loss and all model components are differentiable.
However, when training with just $\ell(\hat Y, Y^*)$ as the loss function, we find that our network rapidly converges to a local minimum where only the reference frame filter $f_1$ is nonzero.
Stochastic gradient descent on our basic loss appears to have difficulty escaping this local minimum, presumably because multi-image alignment and denoising is more difficult than single-image denoising, and because the basic loss does not directly incentivize training to consider anything but the reference frame.
To encourage the network to use the other frames, we use an annealing strategy that initially encourages our filters to individually align and denoise each image in the burst before trying to produce a full 3D filter bank that correctly weights each frame in relation to the others.

Consider the result of applying filters $f_1,\ldots, f_N$ to the frames $X_1,\ldots,X_N$. This yields a stack of $N$ filtered images $f_1(X_1),\ldots,f_N(X_N)$ that can be averaged to produce $\hat Y$.
We add an additional image-space loss against $Y^*$ for each of these intermediate outputs, which is slowly reduced during training. Our final time varying loss is
\begin{equation}
\resizebox{1.0\linewidth}{!}{$
\displaystyle
\mathcal L(X; Y^*, t) = \ell\left(\frac 1 N \sum_{i=1}^N f_i(X_i), Y^*\right) + \beta \alpha^t \sum_{i=1}^N \ell \left( f_i(X_i), Y^* \right).
$}
\label{eq:depth_loss_compositional}
\end{equation}
Here $\beta$ and $0 < \alpha < 1$ are hyperparameters controlling the annealing schedule, and $t$ is the iteration during optimization.
When $\beta \alpha^s \gg 1$, the second term encourages each filter to shift and denoise its corresponding alternate image in the burst independently.
As $t$ approaches $\infty$, this constraint disappears.
In all experiments, we use $\beta=100$ and $\alpha=.9998$, which leads to the second term being phased out around $t=40,000$.
For these values of $\alpha$ and $\beta$, $\mathcal L(\cdot)$ is initially dominated by the second term in Eq.~\ref{eq:depth_loss_compositional}, so annealing can be thought of as a pretraining phase where the KPN is first trained to align and denoise each frame individually before attempting to process the entire burst.

We find that the network's ability to shift alternate frames to correct for misalignment remains intact once the annealed term is essentially zero. After the constraint that each $f_i(X_i)$ should individually resemble ground truth disappears, the network learns to reweight the relative strength of each $f_i$ such that well aligned frames contribute strongly and poorly aligned frames are ignored (see Fig.~\ref{fig:noisevarying}).

We implement our network in Tensorflow~\cite{tensorflow2015} and optimize using Adam~\cite{KingmaB14} with learning rate $10^{-4}$. Our batch size is $4$ and each synthetic burst in the batch has size $128\times 128 \times 8$. We train for one million iterations on an NVIDIA K40 GPU, which takes 4-5 days. At test time, the network can process about 0.7 megapixels/sec on an NVIDIA GTX 1080 Ti.

\section{Experiments}

We first quantitatively assess our method on a synthetic test set, followed by an analysis of its interpretability.
To independently evaluate our design decisions, we conduct a set of ablations and measure the effect of our annealed loss, noise model, and kernel prediction architecture.
Finally, we qualitatively evaluate our model (and demonstrate its ability to generalize) on real bursts captured by a mobile phone and compare against several recent techniques.

We present results on grayscale images because all commonly available real-world linear image data has a Bayer color mosaic. Including demosaicking in our imaging pipeline makes comparison difficult and unfairly biases evaluation against our baseline techniques.
To produce grayscale images from our Bayer raw dataset collected from real cameras, we average each $2\times 2$ Bayer quad into a single pixel and update its noise estimate accordingly.

\begin{table}[]
\begin{center}
\resizebox{3.25in}{!}{
\Huge
\begin{tabular}{ l || cc | cc | cc| cc}
\multicolumn{1}{c}{} & 
\multicolumn{2}{c}{Gain $\propto$ 1} & 
\multicolumn{2}{c}{Gain $\propto$ 2} & 
\multicolumn{2}{c}{Gain $\propto$ 4} & 
\multicolumn{2}{c}{Gain $\propto$ 8} \\
Algorithm & PSNR & SSIM & PSNR & SSIM & PSNR & SSIM & PSNR & SSIM \\
\hline
ref. frame              & 28.70  & 0.733  & 24.19  & 0.559  & 19.80  & 0.372  & 15.76  & 0.212 \\
burst avg.               & 24.70  & 0.628  & 24.06  & 0.552  & 22.58  & 0.431  & 20.00  & 0.285 \\
HDR+\cite{HdrPlus}           & 31.96  & 0.850  & 28.25  & 0.716  & 24.25  & 0.531  & 20.05  & 0.334 \\
BM3D~\cite{BM3D}                  & 33.89  & 0.910  & 31.17  & 0.850  & 28.53  & 0.763  & 25.92  & 0.651 \\
NLM~\cite{NonlocalMeans}                & 33.23  & 0.897  & 30.46  & 0.825  & 27.43  & 0.685  & 23.86  & 0.475 \\
VBM4D~\cite{VBM4D}               & 34.60  & 0.925  & 31.89  & 0.872  & 29.20  & 0.791  & 26.52  & 0.675 \\
\hline 
direct     & 35.93  & 0.948  & 33.36  & 0.910  & 30.70  & 0.846  & \cellcolor{Yellow} 27.97  & \cellcolor{Yellow} 0.748 \\
KPN, 1 frame & 34.95  & 0.932  & 32.07  & 0.878  & 29.22  & 0.791  & 26.29  & 0.657 \\
KPN, no ann.  & 35.42  & 0.944  & 33.01  & 0.903  & 30.46  & 0.836  & 27.65  & 0.724 \\
KPN, $\sigma$ blind  & 36.41  & 0.954  & 33.83  & 0.918  & 30.71  & 0.848  & 22.37  & 0.497 \\
KPN  & \cellcolor{Yellow} 36.47  & \cellcolor{Yellow} 0.955  & \cellcolor{Yellow} 33.93  & \cellcolor{Yellow} 0.920  & \cellcolor{Yellow} 31.19  & \cellcolor{Yellow} 0.857  & 27.97  & 0.741 \\

\end{tabular}
}
\caption{Performance on our linear synthetic test set at various gains (noise levels). Our networks were not trained on the noise levels implied by the gain evaluated in the fourth column (see Fig.~\ref{fig:gains}). 
\label{table:errA}}
\end{center}
\end{table}

\begin{table}[]
\begin{center}
\resizebox{3.25in}{!}{
\Huge
\begin{tabular}{ l || cc | cc | cc | cc}
\multicolumn{1}{c}{} & 
\multicolumn{2}{c}{PSNR $=$ 25} & 
\multicolumn{2}{c}{PSNR $=$ 20} & 
\multicolumn{2}{c}{PSNR $=$ 15} & 
\multicolumn{2}{c}{PSNR $=$ 10} \\
Algorithm & PSNR & SSIM & PSNR & SSIM & PSNR & SSIM & PSNR & SSIM \\
\hline

ref. frame & 25.00 & 0.590 & 
20.00 & 0.377 & 
15.00 & 0.199 & 
10.00 & 0.086\\ 
burst avg. & 25.39 & 0.624 & 
24.10 & 0.502 & 
21.68 & 0.341 & 
18.08 & 0.193\\ 
VBM4D & 32.85 & 0.902 & 
30.00 & 0.826 & 
27.40 & 0.723 & 
25.16 & 0.612\\ 
\hline
direct, $\sigma$ blind & 33.70 & 0.917 & 
28.54 & 0.753 & 
20.76 & 0.418 & 
14.95 & 0.193\\ 
direct & 33.68 & 0.918 & 
30.87 & 0.859 & 
28.06 & \cellcolor{Yellow} 0.771 & 
\cellcolor{Yellow} 25.43 & \cellcolor{Yellow} 0.663\\ 
KPN, $\sigma$ blind & 33.28 & 0.913 & 
25.35 & 0.624 & 
15.87 & 0.227 & 
10.14 & 0.088\\ 
KPN & \cellcolor{Yellow} 33.97 & \cellcolor{Yellow} 0.929 & 
\cellcolor{Yellow} 31.25 & \cellcolor{Yellow} 0.870 & 
\cellcolor{Yellow} 28.07 & 0.758 & 
23.89 & 0.526\\ 
\end{tabular}
}
\caption{Performance on a gamma-corrected version of our synthetic test set with additive white Gaussian noise at four different PSNR levels. The networks without a noise parameter (``$\sigma$ blind'') do not generalize as well to this case, but the networks with a noise parameter generalize well, matching VBM4D's performance in the scenario for which it was designed.
\label{table:errsRGB}}
\end{center}
\end{table}

\newcommand{\synthcompwidthwhole}{.185\textwidth}
\newcommand{\synthcompwidth}{.112\textwidth}

\begin{figure*}[t!]
\begin{center}
\begin{tabular*}{\textwidth}{@{\extracolsep{\fill}}ccccccc@{\extracolsep{\fill}}}
Whole image & Truth & Ref. frame & Average & VBM4D & Direct synth. & KPN\\
\begin{subfigure}[b]{\synthcompwidthwhole}\includegraphics[width=\textwidth]{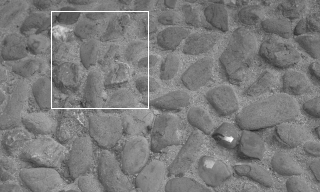} \end{subfigure} & 
\begin{subfigure}[b]{\synthcompwidth}\includegraphics[width=\textwidth]{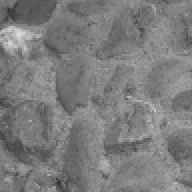} \end{subfigure} & 
\begin{subfigure}[b]{\synthcompwidth}\includegraphics[width=\textwidth]{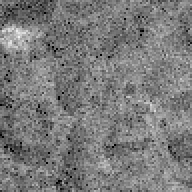} \end{subfigure} & 
\begin{subfigure}[b]{\synthcompwidth}\includegraphics[width=\textwidth]{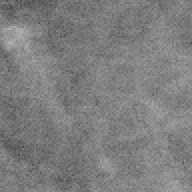} \end{subfigure} & 
\begin{subfigure}[b]{\synthcompwidth}\includegraphics[width=\textwidth]{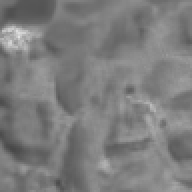} \end{subfigure} & 
\begin{subfigure}[b]{\synthcompwidth}\includegraphics[width=\textwidth]{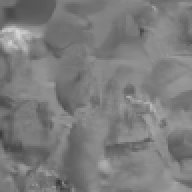} \end{subfigure} & 
\begin{subfigure}[b]{\synthcompwidth}\includegraphics[width=\textwidth]{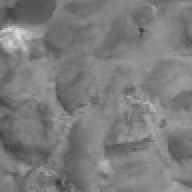} \end{subfigure} \\
\begin{subfigure}[b]{\synthcompwidthwhole}\includegraphics[width=\textwidth]{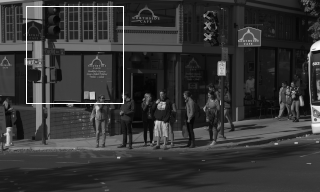} \end{subfigure} & 
\begin{subfigure}[b]{\synthcompwidth}\includegraphics[width=\textwidth]{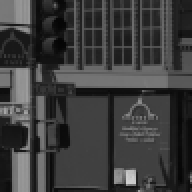} \end{subfigure} & 
\begin{subfigure}[b]{\synthcompwidth}\includegraphics[width=\textwidth]{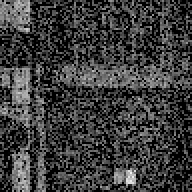} \end{subfigure} & 
\begin{subfigure}[b]{\synthcompwidth}\includegraphics[width=\textwidth]{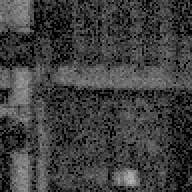} \end{subfigure} & 
\begin{subfigure}[b]{\synthcompwidth}\includegraphics[width=\textwidth]{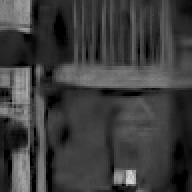} \end{subfigure} & 
\begin{subfigure}[b]{\synthcompwidth}\includegraphics[width=\textwidth]{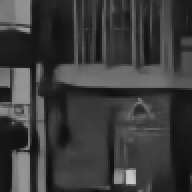} \end{subfigure} & 
\begin{subfigure}[b]{\synthcompwidth}\includegraphics[width=\textwidth]{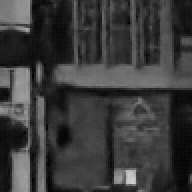} \end{subfigure} \\
\end{tabular*}
\end{center}
\vspace{-5mm}
  \caption{Example results from our synthetic test set. In the top row, we see that VBM4D and the direct synthesis network both produce overly smooth output. In the bottom row, we can see the difference in artifacts produced by each method on an extremely noisy region.}
\label{fig:synthtable}
\end{figure*}

\subsection{Results on synthetic test set}

We report quantitative results on a test set generated with nearly the same procedure as our training set, using 73 linear raw images from a Canon 5D Mark II DSLR to generate image patches instead of our internet images.
These images were taken in bright daylight at low ISO for minimum noise and were deliberately underexposed to avoid clipping highlights.
The maximum image intensity is scaled to 1 after black level subtraction to simulate an exposure that uses the complete dynamic range of the camera.
Misalignment is added with the same procedure as in the training set.

To quantitatively compare to other methods, we evaluate performance at four noise levels corresponding to a fixed set of shot and read noise parameters.
These correspond to ISO settings on a digital camera, where each category is one full photographic ``stop'' higher than the previous (twice the gain or sensitivity to light).
All error metrics (PSNR and SSIM) are computed after gamma correction to better reflect perceptual quality. Results are in Table~\ref{table:errA}.

We evaluate our techniques against several baselines. 
Burst averaging (``burst avg.'') is simply the per-pixel mean of all images in the burst, which performs temporal denoising well but lacks spatial denoising and produces significant errors in the presence of misalignment. ``HDR+'' is the method from~\cite{HdrPlus}, with its spatial denoising disabled by setting $c=0$ in Eq.~7. This method performs similarly to burst averaging but avoids introducing error in the case of misalignment.
Non-local means (NLM)~\cite{NonlocalMeans} and VBM4D~\cite{VBM4D} are multi-frame methods based on finding similar patches and groups of patches across the burst, and BM3D~\cite{BM3D} is a single-frame method based on a similar premise. The non-local means method is implemented with 2D $13 \times 13$ patches found in all of the frames in the burst, accelerated using PCA~\cite{tasdizen2008principal}. 
The ``KPN'' results are our model, which we present alongside a series of ablations: the ``1-frame'' model uses only a single frame as input, the ``no ann'' model uses only our basic loss function with no annealing, and the ``$\sigma$ blind'' model omits the known per-pixel noise as input.
The ``direct'' model is an ablation and extension of our approach, in which we modify our network to directly synthesize denoised pixel values.
Instead of reshaping the $K^2N$ feature vectors into per-pixel kernels, we add an additional 3 convolutional layers.
This architecture produces results similar to the KPN with a comparable amount of computation, but tends to produce oversmoothed results (Fig.~\ref{fig:synthtable}, ``Direct Synth.''), which is favorable only in the highest-noise conditions (Tables~\ref{table:errA} and~\ref{table:errsRGB}).

In Table~\ref{table:errsRGB} we provide an additional experiment in which we assume additive white Gaussian noise.
For this experiment we only evaluate against VBM4D~\cite{VBM4D}, which was the best-performing baseline in our previous experiment (Table~\ref{table:errA}) and is specifically designed for this noise model.
Again, our networks match VBM4D at all noise levels as measured by both PSNR and SSIM.

For all techniques requiring a single input noise level parameter, we performed a sweep and used the value that performed best; see the supplement for details.

\subsection{Predicted kernels}
\label{sec:predictedkernels}

Our network predicts a stack of 2D kernels at each pixel which we visualize in Fig.~\ref{fig:noisevarying}. Despite being trained on patches with synthetically generated translational misalignment, our model learns to robustly reject large scene motions (see Fig.~\ref{fig:filtspatial}).

In our experiments, annealing proved effective in escaping the local minimum that ignores the alternate frames.
Our per-frame loss with $\beta=100$ is a strong constraint and the network is quickly forced to learn a shifted kernel to correct the $\pm 2$ pixel misalignments in the training data.
Once pretrained, reverting to the base frame is no longer viable since averaging already-shifted kernels across the burst yields superior SNR.
The annealing hyperparameter $\alpha$ did not have much impact after shifted kernels have been pretrained.
With our settings, the annealing schedule became effectively zero after only 3-5\% of training iterations.

\newcommand{\mousewidthA}{0.30\linewidth}
\newcommand{\mousewidth}{0.98\linewidth}
\begin{figure}
\begin{center}

  \begin{subfigure}[b]{\mousewidthA}
    \includegraphics[width=\linewidth]{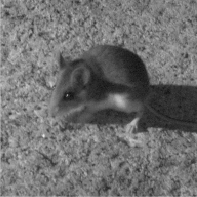}
    \caption{Ref. frame. \label{subfig:mouseA1}}
  \end{subfigure}
  \begin{subfigure}[b]{\mousewidthA}
    \includegraphics[width=\linewidth]{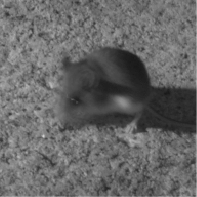}
    \caption{Burst average. \label{subfig:mouseA2}}
  \end{subfigure}
  \begin{subfigure}[b]{\mousewidthA}
    \includegraphics[width=\linewidth]{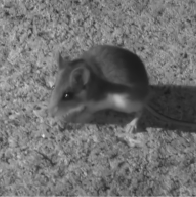}
    \caption{KPN output. \label{subfig:mouseA3}}
  \end{subfigure}
    
  \begin{subfigure}[b]{\mousewidth}
    \includegraphics[width=\linewidth]{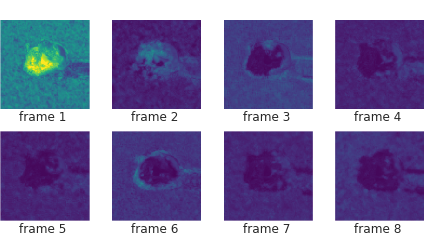}
    \caption{Relative filter strength. \label{subfig:mouseB}}
  \end{subfigure}

\vspace{-0.2in}
\end{center}
  \caption{
  An example of a burst with a sharp reference frame (\ref{subfig:mouseA1}) and a well-aligned static background, but a moving subject. Naive averaging produces a low-noise background and a blurry subject (\ref{subfig:mouseA2}).
  Visualizing the L1 norm of the spatially varying weights allocated to each frame by our predicted filters (\ref{subfig:mouseB}), we see that they draw heavily from the reference frame when denoising the subject, but gather information from multiple frames to produce the background.
  \label{fig:filtspatial}
  }

\vspace{0.2in}

\begin{center}
  \includegraphics[width=1\linewidth]{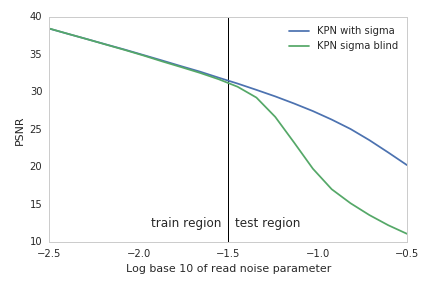}
\end{center}
\vspace{-0.2in}
  \caption{Performance of blind versus noise-aware KPN. The $x$-axis shows the read noise parameter (the shot noise parameter is selected from the gain curve shown in blue in Fig~\ref{fig:gains}). Performance drops off rapidly when using the blind network outside the training region, but the noise-aware network successfully generalizes.
  \label{fig:blind_vs_noise_aware}
  }
\end{figure}

\subsection{Generalization to higher noise levels}
\label{nonblind}

Our network takes as input a per-pixel noise estimate $\sigma'$ together with the images.
One might argue that such ``noise-aware'' algorithms are less useful than ones that can perform ``blind'' denoising without being fed an explicit noise estimate.
In our experiments, including the noise estimate as input only leads to a negligible decrease in training loss (Fig.~\ref{fig:blind_vs_noise_aware}).
However, perhaps surprisingly, we found that including the noise estimate lets our network generalize beyond the noise levels on which it was trained better than the blind variant. 
Fig.~\ref{fig:gains} shows the distribution of noise parameters we sampled at train and test time.
Fig.~\ref{fig:blind_vs_noise_aware} and the final column of Table~\ref{table:errA} demonstrate our performance at noise levels far beyond the training region (note the log scale).
Moreover, Table~\ref{table:errsRGB} shows that our noise-aware method can even denoise gamma-corrected data with additive white Gaussian noise, which was never seen during training.

Beyond generalization, we can treat the noise level input to our noise-aware model as an adjustable parameter $\sigma'$ to tune denoising strength.
Fig.~\ref{fig:noisevarying} shows that the network automatically reweights its filters to incorporate more information from alternate frames as $\sigma'$ increases.

\newcommand{\realcompwidth}{.33\textwidth}

\newcommand{\fivewidth}{0.173\textwidth}
\begin{figure*}
\begin{center}
  \begin{tabular*}{\textwidth}{@{\extracolsep{\fill}}ccccc@{\extracolsep{\fill}}}
  \begin{subfigure}[b]{\fivewidth}
    \includegraphics[width=\textwidth]{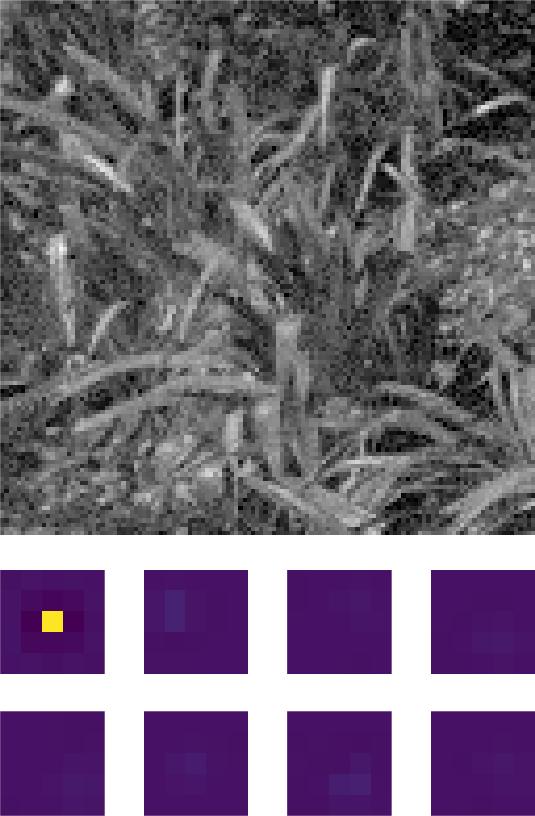}
    \caption{$\sigma' = \hat \sigma_p/4$ \label{subfig:noise_strength1}}
  \end{subfigure}
  &
  \begin{subfigure}[b]{\fivewidth}
    \includegraphics[width=\textwidth]{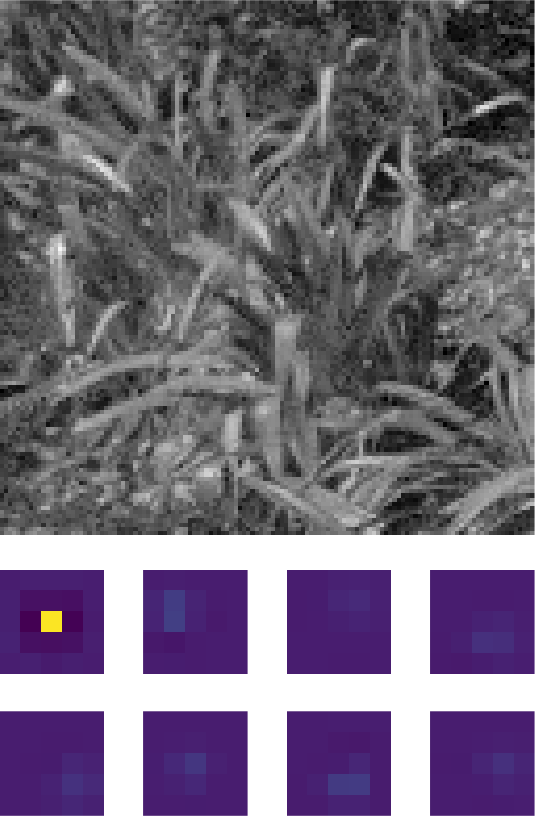}
    \caption{$\sigma' = \hat \sigma_p/2$}
  \end{subfigure}
  &
  \begin{subfigure}[b]{\fivewidth}
    \includegraphics[width=\textwidth]{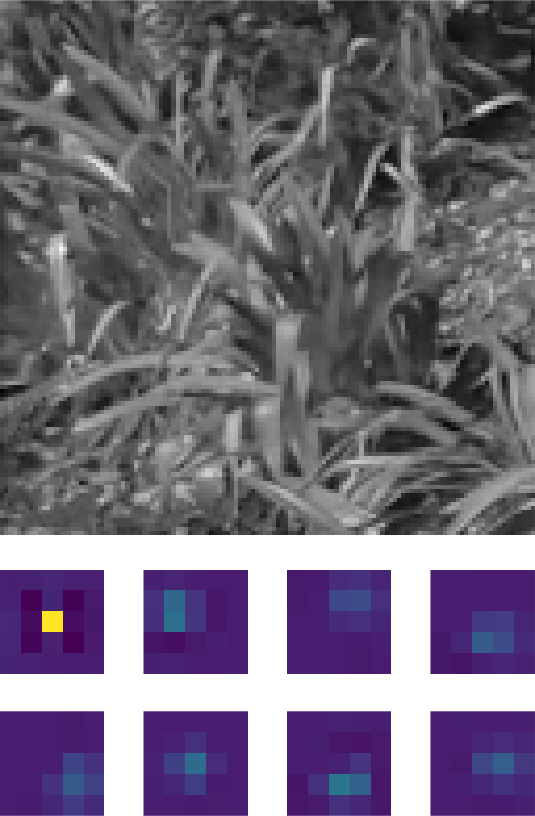}
    \caption{$\sigma' = \hat\sigma_p$}
  \end{subfigure}
  &
  \begin{subfigure}[b]{\fivewidth}
    \includegraphics[width=\textwidth]{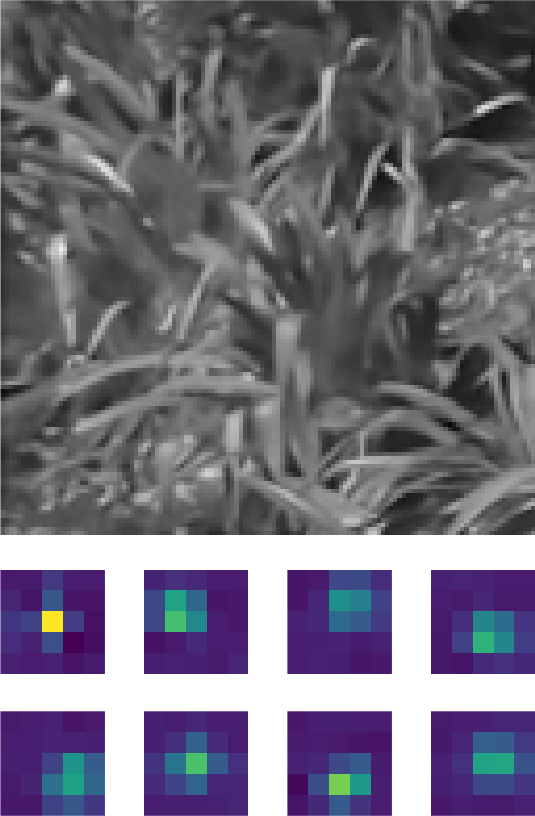}
    \caption{$\sigma' = 2\hat \sigma_p$}
  \end{subfigure}
  &
  \begin{subfigure}[b]{\fivewidth}
    \includegraphics[width=\textwidth]{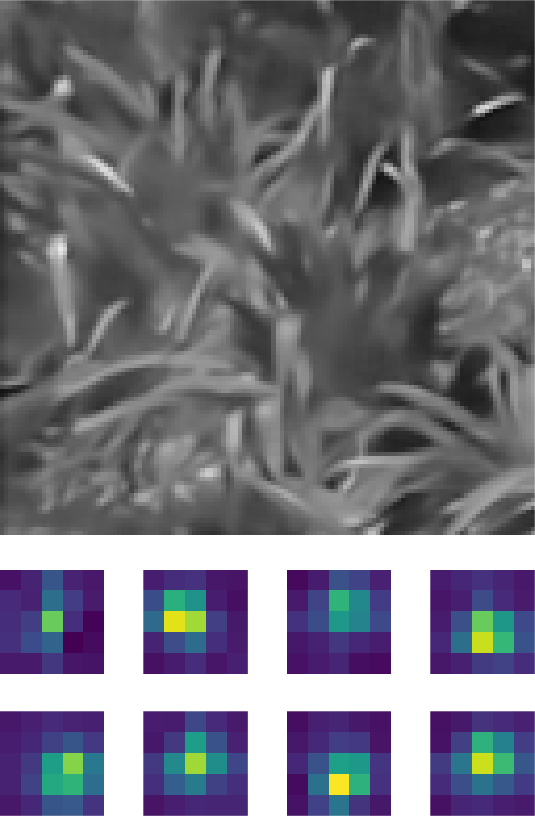}
    \caption{$\sigma' = 4\hat \sigma_p$ \label{subfig:noise_strength6}}
  \end{subfigure}
  \end{tabular*}
\end{center}
\vspace{-0.2in}
  \caption{
  Because our model takes the expected noise level of the image being denoised as input, it is straightforward to analyze its behavior by varying the input noise with a fixed input burst.
  In Figs.~\ref{subfig:noise_strength1} through \ref{subfig:noise_strength6} we pass our KPN model the same input burst images but with differing scalar multiples of the actual estimated noise level $\hat \sigma_p$ (see Eq.~\ref{eq:sigmap}). We visualize the resulting output images (top) and the mean over the two image dimensions of the predicted filter kernels (bottom) for each of the $8$ frames in the burst.
  When the noise level is understated (a-b), the denoising is conservative and the predicted filter stack becomes a delta function on the reference frame, producing an output image identical to the base framet.
  When the noise level is overstated (d-e), the spatial support of the filters widens, the filters for alternate frames strengthen, and the output image becomes smoother.
  \label{fig:noisevarying}
  }

\vspace{0.45in}

\begin{center}

\begin{subfigure}[b]{\realcompwidth}
    \includegraphics[width=\textwidth]{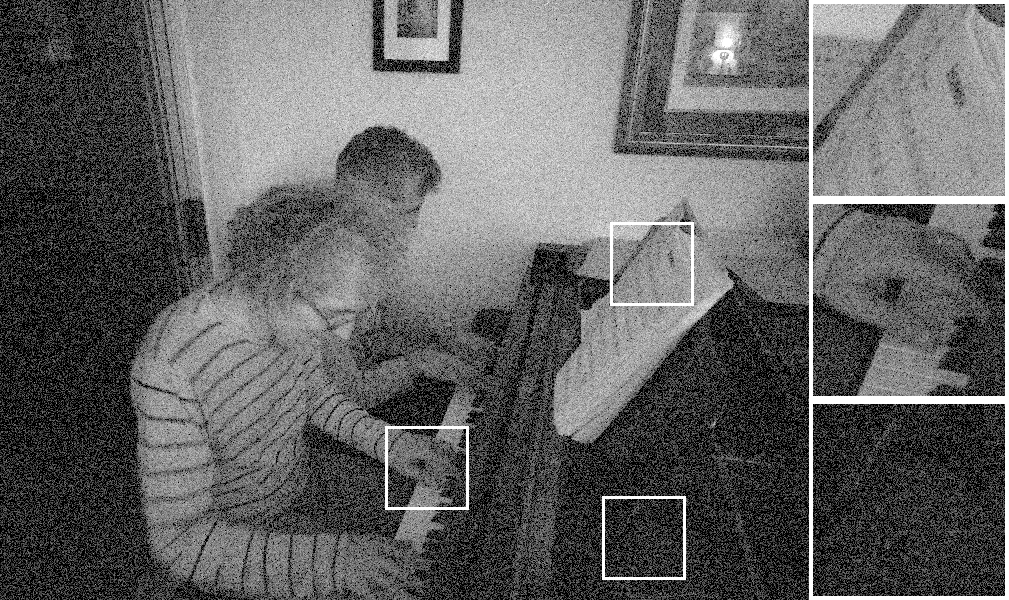}
    \caption{Reference frame \label{subfig:realtable1A}}
\end{subfigure}
\begin{subfigure}[b]{\realcompwidth}
    \includegraphics[width=\textwidth]{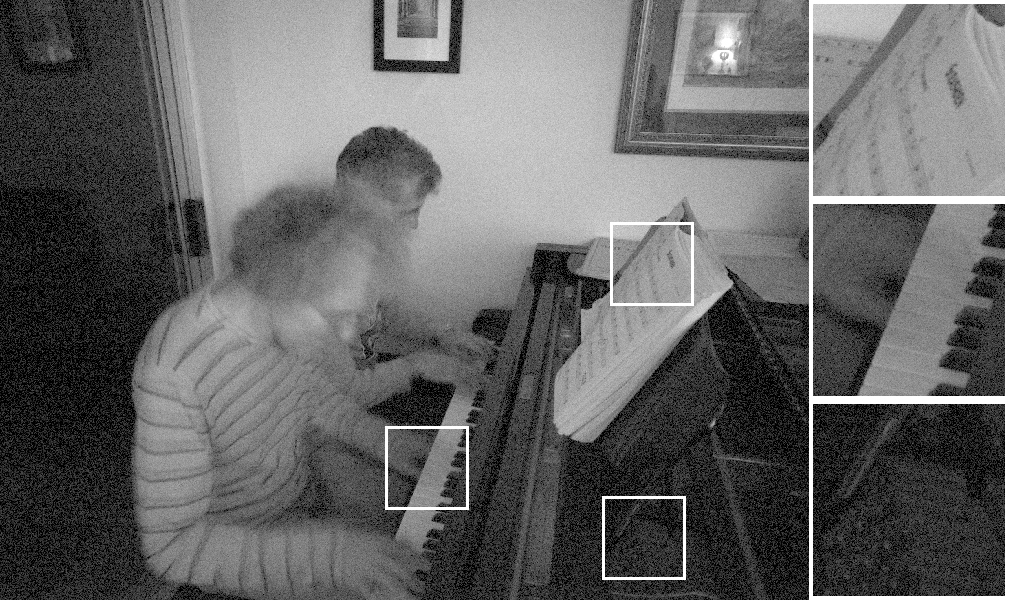}
    \caption{Burst average \label{subfig:realtable1B}}
\end{subfigure}
\begin{subfigure}[b]{\realcompwidth}
    \includegraphics[width=\textwidth]{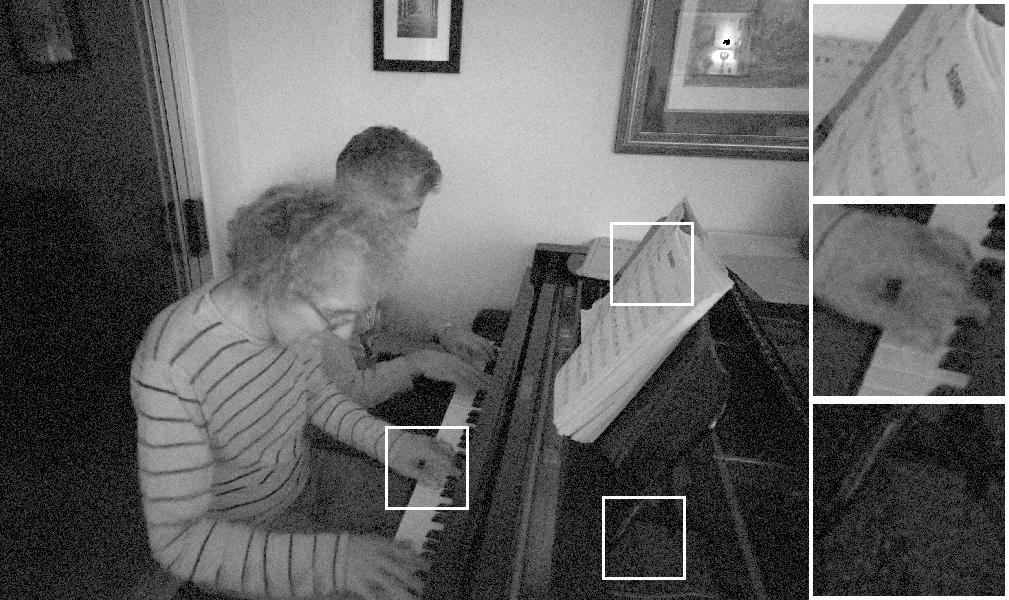}
    \caption{HDR+~\cite{HdrPlus} \label{subfig:realtable1C}}
\end{subfigure}
\begin{subfigure}[b]{\realcompwidth}
    \includegraphics[width=\textwidth]{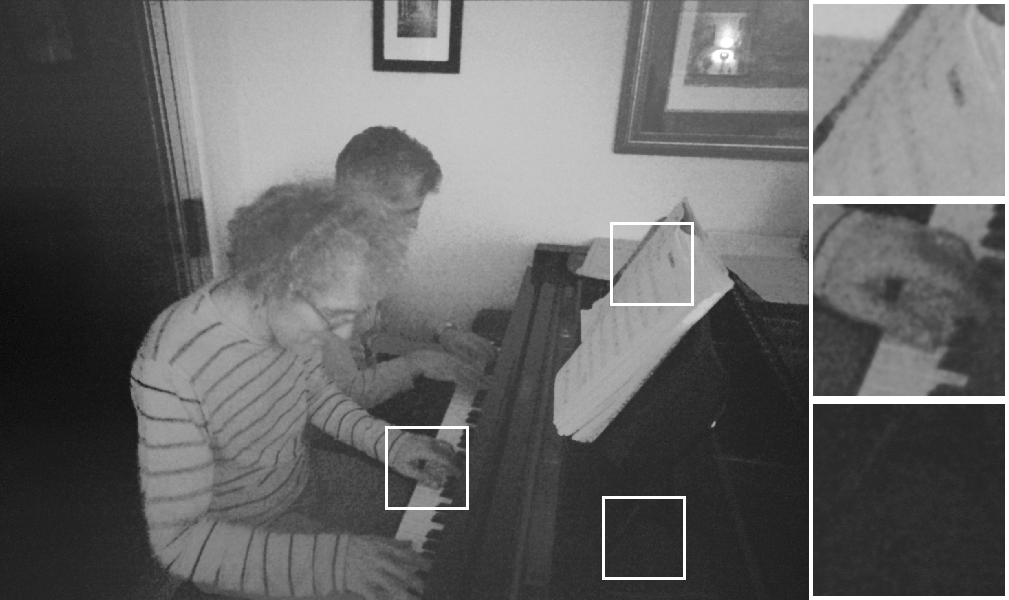}
    \caption{Non-local means~\cite{NonlocalMeans} \label{subfig:realtable1D}}
\end{subfigure}
\begin{subfigure}[b]{\realcompwidth}
    \includegraphics[width=\textwidth]{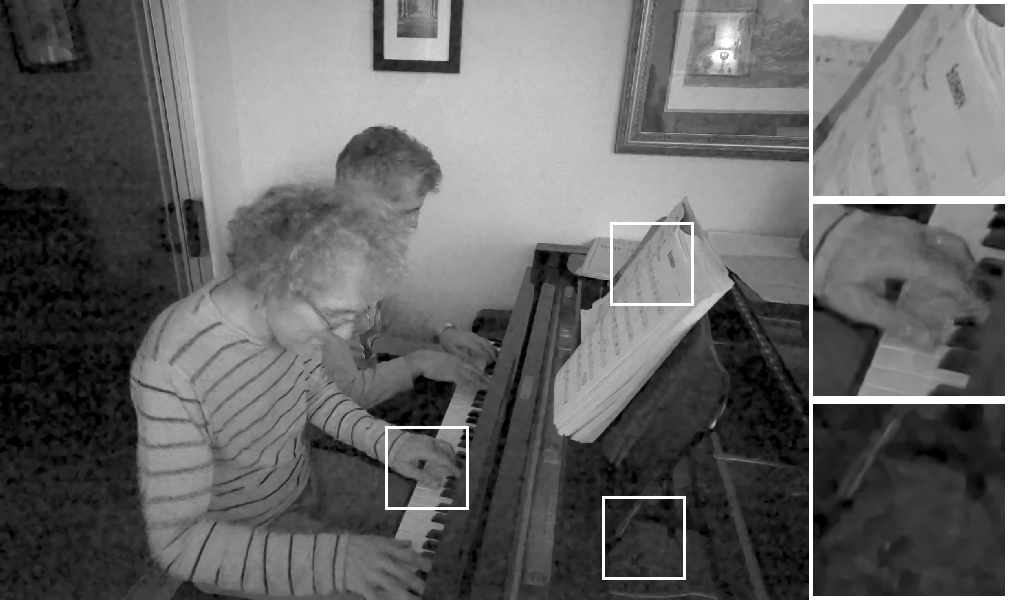}
    \caption{VBM4D~\cite{VBM4D} \label{subfig:realtable1E}}
\end{subfigure}
\begin{subfigure}[b]{\realcompwidth}
    \includegraphics[width=\textwidth]{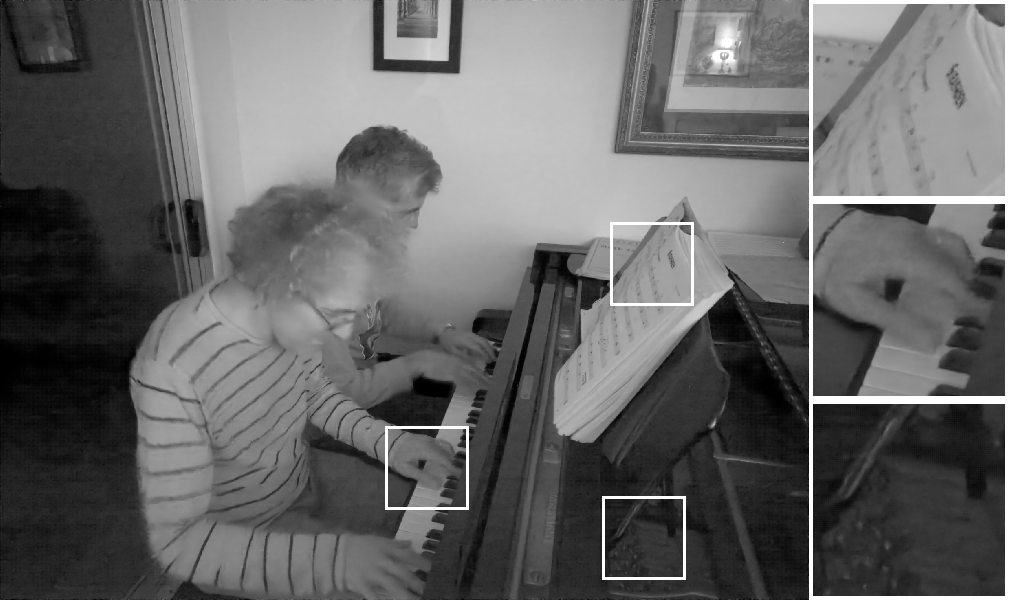}
    \caption{Our KPN model \label{subfig:realtable1F}}
\end{subfigure}
  \caption{Results on a real handheld image burst. While most methods achieve reasonable denoising performance in brighter regions (top inset), both NLM and VBM4D fail on deep shadows (bottom inset). The foreground pianist moves significantly over the course of the burst and simple averaging blurs away details. Conventional techniques that robustly average frames bias the output towards the reference frame but still retain some noise. Our technique (f) recovers the hand (middle inset) while removing more noise than the baseline techniques, without adding artifacts.
  \label{fig:pianoman}
  }
\end{center}
\end{figure*}

\subsection{Generalizing to real data}

We compare our method to several state-of-the-art conventional denoisers on raw bursts captured with a Nexus 6P cellphone in under dim lighting.
We minimally preprocess the burst by subtracting the black level, suppressing hot pixels, and performing a coarse whole-pixel alignment of alternate frames to the reference without resampling, which eliminates some of the globally coherent motion from hand shake but cannot remove scene motion.

Despite having been trained on synthetic data, our method is able to recover detail in the presence of significant noise and does not produce artifacts in the presence of large scene motion.
See Figs.~\ref{fig:realtable1} and \ref{fig:pianoman} for a qualitative comparison between our results and baseline techniques. The supplement contains additional results.

\section{Conclusion}

We have presented a learning-based method for jointly denoising bursts of images captured by handheld cameras.
By synthesizing training data based on a physical image formation model, we are able to train a deep neural network that outperforms the state-of-the-art on both synthetic and real datasets.
A key component to successfully training our kernel prediction network is an annealed loss function based on a heuristic understanding of how kernels handle motion.

\clearpage

{\small
\bibliographystyle{ieee}
\bibliography{egbib}

\begin{thebibliography}{10}\itemsep=-1pt

\bibitem{tensorflow2015}
M.~Abadi, A.~Agarwal, P.~Barham, E.~Brevdo, Z.~Chen, C.~Citro, G.~S. Corrado,
  A.~Davis, J.~Dean, M.~Devin, S.~Ghemawat, I.~Goodfellow, A.~Harp, G.~Irving,
  M.~Isard, Y.~Jia, R.~Jozefowicz, L.~Kaiser, M.~Kudlur, J.~Levenberg,
  D.~Man\'{e}, R.~Monga, S.~Moore, D.~Murray, C.~Olah, M.~Schuster, J.~Shlens,
  B.~Steiner, I.~Sutskever, K.~Talwar, P.~Tucker, V.~Vanhoucke, V.~Vasudevan,
  F.~Vi\'{e}gas, O.~Vinyals, P.~Warden, M.~Wattenberg, M.~Wicke, Y.~Yu, and
  X.~Zheng.
\newblock {TensorFlow}: Large-scale machine learning on heterogeneous systems,
  2015.

\bibitem{KernelPredicting}
S.~Bako, T.~Vogels, B.~Mcwilliams, M.~Meyer, J.~Nov\'{a}K, A.~Harvill, P.~Sen,
  T.~Derose, and F.~Rousselle.
\newblock Kernel-predicting convolutional networks for denoising monte carlo
  renderings.
\newblock {\em SIGGRAPH}, 2017.

\bibitem{NonlocalMeans}
A.~Buades, B.~Coll, and J.~M. Morel.
\newblock A non-local algorithm for image denoising.
\newblock {\em CVPR}, 2005.

\bibitem{BM3D}
K.~Dabov, A.~Foi, V.~Katkovnik, and K.~Egiazarian.
\newblock Image denoising by sparse 3-d transform-domain collaborative
  filtering.
\newblock {\em IEEE Transactions on Image Processing}, 2007.

\bibitem{DynamicFilter}
B.~De~Brabandere, X.~Jia, T.~Tuytelaars, and L.~Van~Gool.
\newblock Dynamic filter networks.
\newblock {\em NIPS}, 2016.

\bibitem{PhysInteract}
C.~Finn, I.~J. Goodfellow, and S.~Levine.
\newblock Unsupervised learning for physical interaction through video
  prediction.
\newblock {\em NIPS}, 2016.

\bibitem{GharbiDemosaic}
M.~Gharbi, G.~Chaurasia, S.~Paris, and F.~Durand.
\newblock Deep joint demosaicking and denoising.
\newblock {\em ACM TOG}, 2016.

\bibitem{HdrPlus}
S.~W. Hasinoff, D.~Sharlet, R.~Geiss, A.~Adams, J.~T. Barron, F.~Kainz,
  J.~Chen, and M.~Levoy.
\newblock Burst photography for high dynamic range and low-light imaging on
  mobile cameras.
\newblock {\em SIGGRAPH Asia}, 2016.

\bibitem{Healey1994}
G.~Healey and R.~Kondepudy.
\newblock Radiometric {CCD} camera calibration and noise estimation.
\newblock {\em TPAMI}, 1994.

\bibitem{Proximal}
F.~Heide, S.~Diamond, M.~Nie{\ss}ner, J.~Ragan-Kelley, and W.~G. Heidrich, W.
\newblock Proximal: Efficient image optimization using proximal algorithms.
\newblock {\em ACM TOG}, 2016.

\bibitem{FlexIsp}
F.~Heide, M.~Steinberger, Y.-T. Tsai, M.~Rouf, D.~Pająk, D.~Reddy, O.~Gallo,
  J.~L. abd Wolfgang~Heidrich, K.~Egiazarian, J.~Kautz, and K.~Pulli.
\newblock {FlexISP}: A flexible camera image processing framework.
\newblock {\em SIGGRAPH Asia}, 2014.

\bibitem{KingmaB14}
D.~P. Kingma and J.~Ba.
\newblock Adam: {A} method for stochastic optimization.
\newblock {\em CoRR}, abs/1412.6980, 2014.

\bibitem{Openimages}
I.~Krasin, T.~Duerig, N.~Alldrin, V.~Ferrari, S.~Abu-El-Haija, A.~Kuznetsova,
  H.~Rom, J.~Uijlings, S.~Popov, A.~Veit, S.~Belongie, V.~Gomes, A.~Gupta,
  C.~Sun, G.~Chechik, D.~Cai, Z.~Feng, D.~Narayanan, and K.~Murphy.
\newblock Openimages: A public dataset for large-scale multi-label and
  multi-class image classification.
\newblock {\em Dataset available from https://github.com/openimages}, 2017.

\bibitem{ReliableMotion}
C.~Liu and W.~T. Freeman.
\newblock A high-quality video denoising algorithm based on reliable motion
  estimation.
\newblock {\em ECCV}, 2010.

\bibitem{VoxelFlow}
Z.~Liu, R.~Yeh, X.~Tang, Y.~Liu, and A.~Agarwala.
\newblock Video frame synthesis using deep voxel flow.
\newblock {\em ICCV}, 2017.

\bibitem{FastBurst}
Z.~Liu, L.~Yuan, X.~Tang, M.~Uyttendaele, and J.~Sun.
\newblock Fast burst images denoising.
\newblock {\em SIGGRAPH Asia}, 2014.

\bibitem{VBM4D}
M.~Maggioni, G.~Boracchi, A.~Foi, and K.~Egiazarian.
\newblock Video denoising, deblocking, and enhancement through separable 4-d
  nonlocal spatiotemporal transforms.
\newblock {\em IEEE Transactions on Image Processing}, 2012.

\bibitem{AdaConv}
S.~Niklaus, L.~Mai, and F.~Liu.
\newblock Video frame interpolation via adaptive convolution.
\newblock {\em CVPR}, 2017.

\bibitem{AdaSepConv}
S.~Niklaus, L.~Mai, and F.~Liu.
\newblock Video frame interpolation via adaptive separable convolution.
\newblock {\em ICCV}, 2017.

\bibitem{PeronaMalik1990}
P.~Perona and J.~Malik.
\newblock Scale-space and edge detection using anisotropic diffusion.
\newblock {\em TPAMI}, 1990.

\bibitem{BenchmarkingDenoising}
T.~Plotz and S.~Roth.
\newblock Benchmarking denoising algorithms with real photographs.
\newblock {\em CVPR}, 2017.

\bibitem{Rudin1992}
L.~I. Rudin, S.~Osher, and E.~Fatemi.
\newblock Nonlinear total variation based noise removal algorithms.
\newblock {\em Phys. D}, 1992.

\bibitem{srgb}
M.~Stokes, M.~Anderson, S.~Chandrasekar, and R.~Motta.
\newblock A standard default color space for the {Internet} --- {sRGB}.
\newblock \url{http://www.color.org/contrib/sRGB.html}, 1996.

\bibitem{DeepDeblurring}
S.~Su, M.~Delbracio, J.~Wang, G.~Sapiro, W.~Heidrich, and O.~Wang.
\newblock Deep video deblurring.
\newblock {\em CoRR}, abs/1611.08387, 2016.

\bibitem{DetailRevealing}
X.~Tao, H.~Gao, R.~Liao, J.~Wang, and J.~Jia.
\newblock Detail-revealing deep video super-resolution.
\newblock {\em ICCV}, 2017.

\bibitem{tasdizen2008principal}
T.~Tasdizen.
\newblock Principal components for non-local means image denoising.
\newblock In {\em ICIP 2008}, pages 1728--1731. IEEE, 2008.

\bibitem{DeepRNNs}
X.~Y. Xinyuan~Chen, Li~Song.
\newblock Deep rnns for video denoising, 2016.

\bibitem{CrossConv}
T.~Xue, J.~Wu, K.~L. Bouman, and W.~T. Freeman.
\newblock Visual dynamics: Probabilistic future frame synthesis via cross
  convolutional networks.
\newblock {\em NIPS}, 2016.

\bibitem{ResidDenoising}
K.~Zhang, W.~Zuo, Y.~Chen, D.~Meng, and L.~Zhang.
\newblock Beyond a gaussian denoiser: Residual learning of deep cnn for image
  denoising.
\newblock {\em IEEE Transactions on Image Processing}, 2017.

\end{thebibliography}
}

\newpage
\appendix

\section{Derivation of noise model parameters}

We can calculate the variance of our output signal as a function of the analog and digital gains we apply to the photoelectron count on the sensor as well as the read noise:
\begin{table}[h!]
    \centering
    \begin{tabular}{lcc}
         & True signal & Variance \\ \hline
        Initial photoe. count & $q$ & $q$ \\
        Analog gain $g_a$ & $g_aq$ & $g_a^2q$ \\
        Readout w/ var. $r^2$ & $g_aq$ & $g_a^2q + r^2$ \\
        Digital gain $g_d$ & $g_dg_aq$ & $g_d^2g_a^2q + g_d^2r^2$
    \end{tabular}
    \label{tab:my_label}
\end{table}

Our measured output is thus $z=g_dg_aq$ with variance $(g_d g_a) z + (g_d r)^2$. In terms of the parameters used in the main text, this gives
\begin{equation}
    \sigma_s = g_dg_a, \quad\quad
    \sigma_r^2 = g_d^2 r^2
\end{equation}
Note that $r$ is fixed but $g_a$ and $g_d$ are controlled by the camera. See~\cite{Healey1994} for more details.

\section{Baseline evaluation details}

\subsection{VBM4D}

For our VBM4D~\cite{VBM4D} comparisons, we estimate a single noise level for each burst as 
\begin{equation}
    \sigma_{rms} = \sqrt{\sum_p \hat \sigma_p^2}
\end{equation}
where $p$ varies over all pixels in the reference frame.

To generate Table 1 in the main text, we ran all methods on a synthetic test set in a linear color space with added shot and read noise. To evaluate VBM4D~\cite{VBM4D} fairly, we tried running it both on linear data as-is and on the data after additionally applying gamma correction. We found the final gamma-corrected PSNR to be better after running on the original linear data, likely because of the loss of information from clipping negative values when applying gamma correction before denoising. (We also found this to be the case when running single image BM3D~\cite{BM3D} on the reference frame only.)

We additionally ran a parameter sweep over a multiplier for $\sigma_{rms}$, testing VBM4D with noise parameter $k\sigma_{rms}$ for $k \in [.5, 1, 2, 3]$, with $k=1$ producing the best results.

To generate Table 2, we used a test set in a gamma corrected color space with added constant variance Gaussian noise. VBM4D is designed to work best in this setting, so we did not try any variations. 

\subsection{Nonlocal means}

Nonlocal means~\cite{NonlocalMeans} takes a single noise parameter. For the NLM comparisons, we run a parameter sweep over $k\sigma_{rms}$ with $k \in [.5, 1, 2, 3, 4]$. We find that $k=2$ produces the best results.

\subsection{HDR+}

The merge technique in HDR+ takes a single free parameter $c$ (see equation 7 in~\cite{HdrPlus}). We find that $c\approx 10^{2.5}$ works best on average for our linear data in $[0,1]$.

\end{document}